\documentclass[twoside,11pt]{article}

\usepackage[abbrvbib]{jmlr2e}

\usepackage{jmlr2e}
\usepackage{xcolor}
\usepackage{xspace}
\usepackage{makecell}
\usepackage{booktabs}
\usepackage{listings}
\newcommand*\samethanks[1][\value{footnote}]{\footnotemark[#1]}
\definecolor{dkred}{rgb}{0.5,0,0}
\definecolor{dkgreen}{rgb}{0,0.6,0}
\definecolor{gray}{rgb}{0.5,0.5,0.5}
\definecolor{mauve}{rgb}{0.58,0,0.82}

\newcommand{\system}{{PyGOD}\xspace}

\hypersetup{hidelinks}

\usepackage{lastpage}
\jmlrheading{25}{2024}{1-\pageref{LastPage}}{7/23; Revised
2/24}{5/24}{23-0963}{Kay Liu, Yingtong Dou, Xueying Ding, Xiyang Hu, Ruitong Zhang, Hao Peng, Lichao Sun, Philip S. Yu}

\ShortHeadings{PyGOD: A Python Library for Graph Outlier Detection}
{Liu, Dou, Ding, Hu, Zhang, Peng, Sun, and Yu}
\firstpageno{1}

\begin{document}

\title{PyGOD: A Python Library for Graph Outlier Detection}

\author{\name Kay Liu$^1$\thanks{Equal contribution. The work was done when Yingtong Dou was at UIC.} \email zliu234@uic.edu \\
       \name Yingtong Dou$^1$$^{,}$$^2$\samethanks \email yidou@visa.com \\
       % \name Yue Zhao$^3$$^{,}$$^4$\samethanks \email yzhao010@usc.edu \\
       \name Xueying Ding$^3$ \email xding2@andrew.cmu.edu \\
       \name Xiyang Hu$^4$ \email xiyanghu@asu.edu \\
       \name Ruitong Zhang$^5$ \email zhangruitong.zrt@alibaba-inc.com \\
       % \name Kaize Ding$^{5}$ \email kaize.ding@northwestern.edu\\
       % \name Canyu Chen$^6$ \email cchen151@hawk.iit.edu \\
       \name Hao Peng$^6$$^{,}$$^{7}$ \email penghcs@gmail.com\\
       % \name Kai Shu$^6$ \email kshu@iit.edu\\
       \name Lichao Sun$^{8}$ \email lis221@lehigh.edu\\
       % \name Jundong Li$^{12}$ \email jundong@virginia.edu\\
       % \name George H. Chen$^4$ \email georgechen@cmu.edu\\
       % \name Zhihao Jia$^4$ \email zhihao@cmu.edu\\
       \name Philip S. Yu$^1$ \email psyu@uic.edu\\
       \addr $^1$University of Illinois Chicago,
       \addr $^2$Visa Research,
       % \addr $^3$University of Southern California,
       \addr $^3$Carnegie Mellon University,
       \addr $^4$Arizona State University,
       \addr $^5$Alibaba Group,
       % \addr $^5$Northwestern University,
       % \addr $^6$Illinois Institute of Technology,
       \addr $^6$Kunming University of Science and Technology,
       \addr $^{7}$Shantou University,
       \addr $^{8}$Lehigh University
       % \addr $^{12}$University of Virginia,
       }

\editor{Sebastian Schelter}

\maketitle

\begin{abstract}%   <- trailing '%' for backward compatibility of .sty file
\system is an open-source Python library for detecting outliers in graph data.
As the first comprehensive library of its kind, \system supports a wide array of leading graph-based methods for outlier detection under an easy-to-use, well-documented API designed for use by both researchers and practitioners. \system provides modularized components of the different detectors implemented so that users can easily customize each detector for their purposes. To ease the construction of detection workflows, \system offers numerous commonly used utility functions.
To scale computation to large graphs, \system supports functionalities for deep models such as sampling and mini-batch processing.
\system uses best practices in fostering code reliability and maintainability, including unit testing, continuous integration, and code coverage.
To facilitate accessibility, \system is released under a BSD 2-Clause license at \url{https://pygod.org} and at the Python Package Index (PyPI).
\end{abstract}


\section{Introduction}
Outlier detection (OD), also known as anomaly detection, is a key machine learning task to identify deviant samples from the general data distribution \citep{aggarwal2017introduction,li2022ecod}.
With the increasing importance of graph data in both research and real-world applications \citep{ding2021few,huang2021therapeutics,fu2021mimosa,zhou2021subtractive,xu2022contrastive}, detecting outliers with graph-based methods, particularly graph neural networks (GNNs), has recently garnered considerable attention \citep{ma2021comprehensive,ding2019interactive,ding2021inductive,liu2022bond} with many applications such as detecting suspicious activities in social networks \citep{sun2022adversarial,dou2020enhancing} and security systems \citep{cai2021structural}. 

Although there is a long list of libraries for detecting outliers in tabular and time-series data in multiple programming languages (e.g., PyOD \citep{zhao2019pyod}, SUOD \citep{zhao2021suod}, PyODDS \citep{li2020pyodds}, ELKI \citep{Achtert2010visual}, OutlierDetection.jl \citep{muhr2022outlierdetection},  PyTOD \citep{zhao2021tod}, TODS \citep{lai2021tods}, Telemanom \citep{hundman2018detecting}), there is no specialized library for graph outlier detection.

\begin{table}[!t]
\centering
	\label{table:algorithms} % is used to refer this table in the text
    \resizebox{0.9\textwidth}{!}{
	\begin{tabular}{l|ccccr} % centered columns (4 columns)
	    \toprule
		\textbf{Algorithm} &
		\textbf{Backbone} & \textbf{GPU} & \textbf{Sampling} & \textbf{Inductive} & \textbf{Reference} \\
		\midrule
        SCAN & Clustering & No & No & No & \citep{xu2007scan}\\
        GAE & GNN+AE & Yes & Yes & Yes & \citep{kipf2016variational} \\
        Radar & MF & Yes & No & No & \citep{li2017radar} \\
        ANOMALOUS & MF & Yes & No & No & \citep{peng2018anomalous} \\
        ONE & MF & Yes & No & No & \citep{bandyopadhyay2019outlier}\\
        DOMINANT & GNN+AE & Yes & Yes & Yes & \citep{ding2019deep} \\
        DONE & GNN+AE & Yes & Yes & No & \citep{bandyopadhyay2020outlier} \\
        AdONE & GNN+AE & Yes & Yes & No & \citep{bandyopadhyay2020outlier} \\
        AnomalyDAE & GNN+AE & Yes & Yes & Yes & \citep{fan2020anomalydae} \\
        GAAN & GAN & Yes & Yes & Yes  & \citep{chen2020generative}\\
        DMGD & GNN+AE & Yes & Yes & No & \citep{bandyopadhyay2020integrating} \\
        OCGNN & GNN & Yes & Yes & Yes  & \citep{wang2021one} \\
        CoLA & GNN+AE+SSL & Yes & Yes  & Yes & \citep{liu2021anomaly} \\
        GUIDE & GNN+AE & Yes & Yes & Yes  & \citep{yuan2021higher} \\
        CONAD & GNN+AE+SSL & Yes & Yes  & Yes & \citep{xu2022contrastive}\\
        GAD-NR & GNN+AE & Yes & Yes & Yes & \citep{roy2023gadnr} \\
		\bottomrule
	\end{tabular}}
 \caption{Implemented graph outlier detectors in \system v1.1.0.}% title of Table
\end{table}

To bridge this gap, we design the first comprehensive \textbf{Py}thon \textbf{G}raph \textbf{O}utlier \textbf{D}etection library called \system, 
with a couple of key technical advancements and contributions. 
First, it covers a wide array of algorithms with various backbones, including clustering, matrix factorization (MF), generative adversarial networks (GANs), autoencoders (AEs), GNNs, and self-supervised learning (SSL).
\system already supports more than fifteen representative algorithms as shown in Table \ref{table:algorithms}.
Second, \system implements these detection models with a unified API so that the user only needs to prepare the data in a predefined graph format, at which point all outlier detectors in \system can process the data.
Third, \system offers flexible and modularized components of the different outlier detectors implemented, enabling users to customize these detectors according to individual needs. Morover, \system provides many commonly used utility functions to ease the construction of graph outlier detection workflows.
Fourth, \system can scale outlier detection to large graphs using sampling and mini-batch processing.
% Lastly, \system comes with detailed API documentation and examples for clarity and ease of use.
With a focus on code clarity and quality, we provide comprehensive API documentation and examples. Additionally, we provide unit tests with cross-platform continuous integration along with code coverage and maintainability checks.

\begin{figure*}[t]
    \centering
    \resizebox{0.9\textwidth}{!}{
	\includegraphics[width=\linewidth]{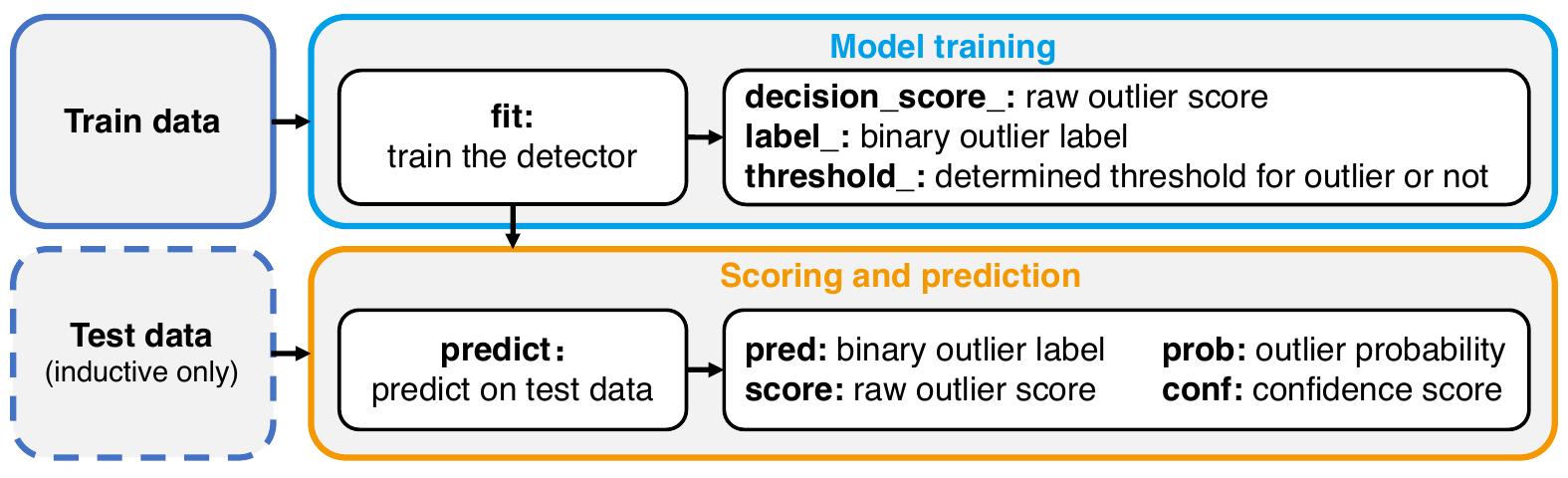}
    }%\vspace{-1em}
	\caption{Demonstration of \system's unified API design.}
	\label{fig:vis}
\end{figure*} 

\section{Library Design and Implementation}
\textbf{Dependency}. \system builds for Python 3.8+ and depends on the popular \texttt{PyTorch} \citep{paszke2019pytorch} and \texttt{PyTorch Geometric (PyG)} \citep{Fey/Lenssen/2019} packages for graph learning on both CPUs and GPUs.
Additionally, \system uses \texttt{NumPy} \citep{harris2020array}, \texttt{SciPy} \citep{virtanen2020scipy}, and \texttt{scikit-learn} \citep{Pedregosa2011scikit}, and \texttt{NetworkX} \citep{SciPyProceedings_11} for data manipulation.

\noindent \textbf{API Design}. As shown in Figure \ref{fig:vis}, inspired by the API design of \texttt{scikit-learn} \citep{sklearn_api} and \texttt{PyOD} \citep{zhao2019pyod}, all detection algorithms in \system inherit from a base class with the same API interface:
(\textit{i}) \texttt{fit} trains the detector, gets the outlier scores (higher means more outlying) and outlier prediction on the input data, and generates necessary statistics for prediction (in the inductive setting); (\textit{ii}) \texttt{predict} leverages the trained model to predict on the input data in the inductive setting (no input data are required in the transductive setting). \texttt{predict} returns a binary prediction (0 for normal samples and 1 for outliers), a raw outlier score, a probability of a sample being an outlier (using the method by \citet{kriegel2011interpreting}), and a confidence score \citep{perini2020quantifying} based on users' needs.
The usage of the above APIs is demonstrated in Code Demo~\ref{lst:label}.

\vspace{0.1in}

\begin{minipage}{0.95\linewidth}
\renewcommand{\lstlistingname}{Code Demo} % Listing->Code
\begin{lstlisting}[caption={Using DOMINANT \citep{ding2019deep} on Cora \citep{Morris+2020}. 
% \textcolor{blue}{import the Planetoid at the top. Should we go through each line of the code in the body text?} \textcolor{red}{I updated the caption; maybe it is fine since cora is quite popular?}
},captionpos=b, label={lst:label}, numbers=left, xleftmargin=0.5em,frame=single,framexleftmargin=1.3em]
   from pygod.utils import load_data                   # import data function
   data = load_data("inj_cora")                        # load built-in dataset

   from pygod.detector import DOMINANT                 # import the detector
   model = DOMINANT(num_layers=4)                      # initialize the detector
   model.fit(data)                                     # train with data
  
   pred, score = model.predict(data,                   # predict labels by default
                                  return_score=True)   # and raw outlier scores
  
   from pygod.metric import eval_roc_auc, eval_f1      # import the metric
   eval_f1(data.y.bool(), pred)                        # evaluate by F1
   eval_roc_auc(data.y.bool(), score)                  # evaluate by AUC
\end{lstlisting}
\end{minipage}

\noindent \textbf{Streamlined Graph Learning with PyG}. We choose to develop \system on top of the popular \texttt{PyG} library for multiple reasons. First, this reduces the complexity of processing graph data for users.
% with a common, unified data format. 
That is, \system only requires the input data to be in the standard graph data format in PyG\footnote{PyG data object: \url{https://pytorch-geometric.readthedocs.io/en/latest/modules/data.html}}.
% Notably, different detection models need distinct information from a \texttt{PyG} graph.
% % , and \system further processes the data for different graph detection models.
% Within the implementation of each detection model, we design an abstract \texttt{process\_graph} method to extract necessary information, e.g., the adjacency matrix, node, and edge attributes, etc., for the underlying detection algorithm. 
Second, most of the detectors use GNNs \citep{kipf2016semi} as their backbone (see Table \ref{table:algorithms}), where PyG already provides an optimized implementation.
Third, PyG is the most popular GNN library with advanced functions like graph sampling and distributed training. Under the PyG framework, we implement mini-batch processing and/or sampling for selected models to accommodate learning with large graphs as shown in Table \ref{table:algorithms}. 

\noindent \textbf{Modularized components and helpful utility functions}. To minimize code redundancy and improve reusability, \system employs modularization in the implementation of deep detectors, dividing different components into \texttt{nn.conv}, \texttt{nn.encoder}, \texttt{nn.decoder}, and \texttt{nn.functional}.
Additionally, a set of helpful utility functions is designed to facilitate graph outlier detection.
In terms of tasks, \system includes \texttt{utils.to\_edge\_score} and \texttt{utils.to\_graph\_score} to enable the adaptation of any node level model to edge level and (sub)graph level outlier detection.
In terms of evaluation, \system offers common metrics for graph outlier detection in the \texttt{metric} module.
In terms of data, \system provides built-in example data sets through \texttt{utils.load\_data}.
Moreover, it offers outlier generator methods in the \texttt{generator} module for injecting both contextual and structural outliers \citep{ding2019deep}. This serves as a solution for model evaluation and benchmarking. 
For more details, please refer to \system documentation\footnote{\label{note2}Documentation: \url{https://docs.pygod.org/}}.

% \textcolor{blue}{add a paragraph to introduce which models are implemented by us and which models are modified based on original source code, add Github references to Table~\ref{table:algorithms}?} \textcolor{red}{Maybe this can be skipped since we need to do updates anyway.}

% \vspace{-0.05in}

\section{Library Robustness and Accessibility}

\textbf{Robustness and Quality}.
While building \system, we follow the best practices of system design and software development.
First, we leverage the continuous integration by \textit{GitHub Actions}\footnote{Continuous integration by GitHub Actions: \url{https://github.com/pygod-team/pygod/actions}} to automate the testing process under various Python versions and operating systems. 
In addition to the scheduled daily test, both commits and pull requests trigger the unit testing.
Notably, we enforce all code to have over 99\% coverage\footnote{Code coverage by Coveralls: \url{https://coveralls.io/github/pygod-team/pygod}}.
By following the \texttt{PEP8} standard, we enforce a consistent coding style and naming convention, which facilitates community collaboration and code readability.  

\noindent \textbf{Accessibility and Community}.
\system comes with detailed API documentation rendered by \texttt{Read the Docs}.
The documentation includes an installation guide as well as interactive examples in  \texttt{Jupyter notebooks}.
To facilitate community contribution, the project is hosted on \texttt{GitHub} with a friendly contribution guide and issue reporting mechanism. At the time of publishing, PyGOD has been widely used in numerous real-world applications including Twitter bot detection \citep{feng2022twibot} and financial fraud detection \citep{huang2022dgraph}, with more than 1,200 GitHub stars and 20,000 PyPI downloads.

% \vspace{-0.05in}

\section{Conclusion and Future Plans}

In this paper, we present the first comprehensive library for graph outlier detection, called \system. \system supports a wide range of detection algorithms with a unified API, rich documentation, and robust code design. These features make it valuable for both academic research and industry applications.
The development plan of \system will focus on multiple aspects: 
% (\textit{i}) including more algorithms for different sub-tasks, e.g., outlier detection in edges and sub-graphs;
% (\textit{i}) collaborating with industries to make \system more practical and tailor the needs of practitioners;
(\textit{i}) enabling detectors to acquire domain knowledge by incorporating different amounts of supervision signals;
(\textit{ii}) optimizing its scalability with the latest advancement in graphs  \citep{jia2020improving}; and 
(\textit{iii}) incorporating automated machine learning to enable intelligent model selection and hyperparameter tuning \citep{zhao2021automatic}.

\clearpage
\newpage

\acks
The authors who are affiliated with the University of Illinois Chicago are supported in part by NSF under grant III-2106758 and POSE-2346158. %G.H.C.~was supported by NSF CAREER award \#2047981. 
%Peng was supported by NSFC under grant 62322202. 
Philip S. Yu and Hao Peng are the corresponding authors.
% (we will add it back once accepted)

\vskip 0.2in
\bibliography{sample}

\end{document}